# SEQUENTIAL ANATOMY LOCALIZATION IN FETAL ECHOCARDIOGRAPHY VIDEO

*Arijit Patra, J A Noble*

Institute of Biomedical Engineering, University of Oxford

## ABSTRACT

Fetal heart motion is an important diagnostic indicator for structural detection and functional assessment of congenital heart disease. We propose an approach towards integrating deep convolutional and recurrent architectures that utilize localized spatial and temporal features of different anatomical substructures within a global spatiotemporal context for interpretation of fetal echocardiography videos. We formulate our task as a cardiac structure localization problem with convolutional architectures for aggregating global spatial context and detecting anatomical structures on spatial region proposals. This information is aggregated temporally by recurrent architectures to quantify the progressive motion patterns. We experimentally show that the resulting architecture combines anatomical landmark detection at the frame-level over multiple video sequences- with temporal progress of the associated anatomical motions to encode local spatiotemporal fetal heart dynamics and is validated on a real-world clinical dataset.

## 1. INTRODUCTION

Fetal ultrasound is a key aspect of pre-natal health assessment across pregnancies worldwide due to its non-invasive nature, relatively low cost and portability. As such, different aspects of physiological assessment of the developing fetus relies upon an understanding of fetal ultrasound videos. Since a key driver for infant mortality worldwide is congenital heart disease, analyzing fetal cardiac videos is important. Fetal heart anomalies are often misdiagnosed or missed due to a lack of equipment and expertise, which leads to many health systems leaving out fetal cardiac screening from the mandatory requirements of 20-week abnormality scans. Accurate analysis of fetal heart US videos is a difficult task for human sonographers due to indistinct appearances of multiple anatomical structures in a small area. In addition, there exist speckle, shadowing, enhancement and variations in contrast levels in clinical US images and videos. The sonographer often has to perform multiple activities during an US scan such as viewing plane identification, anomaly detection, sex determination and so on. In addition to assessments of structural abnormalities, a regionalized functional analysis is vital to overall diagnoses of fetal cardiac abnormalities. This includes an analysis of motion patterns of anatomical structures like the valves, the formane ovule and the ventricle walls, with respect to global heart motion, as symptoms of several cardiac abnormalities may lie in anomalous motion patterns of these structures. Thus, a local analysis of such motions in a global context of the cardiac cycle is vital. Here, we propose a convolutional-recurrent pipeline to detect multiple anatomical structures in the fetal heart from echocardiography videos and quantify their motion extents in an end-systole to end-diastole cycle.

Deep learning techniques have evolved to become state-of-the-art algorithms for different aspects of image, video and language processing, ranging from image classification and object detection to representations for tasks in spatiotemporal domains like video classification and localization. While strides have been made to understand videos of natural scenes, our ultrasound based task is challenged by the requirements to detect spatial structures and action progress for multiple structures represented by multiple detection candidates. Prior work related to action localization primarily dealt with RGB videos with strong visual and temporal cues. In contrast, feature definition in ultrasound image sequences can be weak and obscured due to speckle, acoustic shadowing and other artefacts. Similar issues render optical flow based methods for temporal modelling sub-optimal for ultrasound videos. Thus, multi-object detection in ultrasound videos, as is the nature of the task of simultaneously localizing anatomical structures here, requires superior methods to integrate spatial and temporal linkages of salient features.

We approach the fetal echocardiography video anatomy localization problem as a 3-step pipeline by first detecting the presence of the heart, followed by detection of anatomical markers at the spatial level incorporating global context. The third step performs a temporal aggregation of relevant structures identified at the frame-level and learns to regress to motion progress by an RNN architecture. This results in identification of local anatomies over multiple frames, creating local 'tubes' of anatomical motion patterns. We term them 'echo tubes' or 'motion tubes' (an 'echo tube' is defined as the interval from the start to end of a motion period of an anatomical structure including all patches in between). To our knowledge this is a novel approach to spatiotemporal ultrasound feature description that extends frame-level multi-object localization and combines it with local cardiac phase information. Our contributions are:

- Learning based detection methods for localization of multiple anatomical structures across frames of fetal echocardiography videos considering as exemplars of the approach, five diagnostically relevant sub-structures.
- Progressive modelling of anatomical motion for detected structures by aggregation of spatial information with temporal regression for local motion progress.
- Quantification of motion for a number of key fetal heart anatomical land-marks within the global context of the fetal cardiac cycle, while correcting for non-deterministic motions due to global fetal motions and probe movement.

## 2. PRIOR WORK

Modelling of spatiotemporal dynamics in vision has been attempted in action recognition domains. [3,4,5] implement video classification through 3D CNNs, two-stream CNNs fusing optical flow and spatial context and CNN-RNN hybrids extending spatial features across temporal domains respectively. While most video classification methods assume a single class per video sequence, action localization has been a challenging problem due to multiple detection, classification and localization tasks in variable spatial and temporal scales. Recently [6] presented a tube proposal network from 3D feature maps leveraging 3D feature map generations from 3D CNNs. Most localization work focused on integrating frame level object detection [1] with methods to build linkages across frames and defining the start and conclusion of disparate actions [7]. Action progress estimation was reported with global context and region proposals coupled with an RNN to regress to action end-points [8] with a single action per class. In ultrasound imaging, [9, 10] leveraged the spatiotemporal dynamics in videos, by characterizing the visibility, viewing planes and localization of the fetal heart using convolutional temporal fusion. To date, there have been no studies that performed anatomical detection with multiple regions of interest with temporal extensions for fetal ultrasound despite the importance of diagnostic assessment of physiological motion observed by medical imaging systems. Video understanding at the local level has largely been untouched for ultrasound where optical flow is difficult to estimate and generating unique motion tubes per structure is non-trivial. Unlike most problems in computer vision, relying on a presence of specific spatial cues is insufficient to isolate distinct motion tubes or to quantify start or end-points. This is because relevant anatomical structures are spatially present throughout but given the periodicity of anatomical motion, the characteristic marker for motion completion involves isolating a single periodic cycle that may vary for different anatomical structures within the fetal heart and with overall heart rates for unhealthy instances.

### 3. SPATIOTEMPORAL ANATOMY LOCALIZATION

Multi-anatomy localization requires the ability to extract multiple region features from a given frame and classify those into the classes of anatomies under study or background. Aggregation of temporal context is then a replication of this process over sequential frames and combining similarly predicted region proposals as an anatomical class in the video. The structure detection is followed by motion quantification for which the anatomical deformation is treated as an action, or a sequence of actions in the physiological environment being visually interpreted. Thus, anatomical movement can be approximated as a constrained action recognition with the periodicity of physiological deformation serving as a marker of progress. Specific to our task of interpreting fetal echocardiography videos, we consider five key anatomical motions, based on their clinical relevance towards diagnosis of 4C view Congenital Heart Disease [9] – the mitral and the tricuspid valves, the left and the right ventricle walls, and the foramen ovule. Initially, the presence of the heart in the four-chamber view is ensured using a pre-trained modified VGG-16 convolutional neural network [2], hereafter referred to as Model-1 that classifies the presence or absence of the 4C view. If the 4C view is apparent in a frame, we proceed with a detection routine that looks for spatial locations of the structures whose motion is of interest. To that end, we adopt a Faster RCNN [1] derived model based on a VGG-16 base augmented by LSTM layers to temporally classify our five structures of interest and background. Extending to temporal localization and motion phase quantification require structures of interest to be recognized across multiple frames and their motion seen as successive detection of same structures over time, with a scheme to impose bounds for start and end of a motion phase.

We follow [1] and generate multiple aspect ratio region proposals with the difference that we generate region proposals directly at input stage and constrain them to be in either 50 x 50 or 50 x 100 patches without overlap to keep computational requirements manageable. An unconstrained generation of bounding boxes with overlapping can potentially lead to a significant redundancy in terms of the probability of finding a structure of interest as each structure appears uniquely only once in a given frame. We compute their class confidence scores (representative of the anatomical structure) and refine the bounding boxes using Model-2, which is a modified RCNN approach. Model-2 also includes LSTM layers to aggregate temporal information over sequential frames. A difference from R-CNN methods is that we do away with 'objectness' calculations and treat backgrounds as a separate class. This is due to the difference of our task in the much lower number of probable classes per region and requirement to pass fully connected layers of patches of the same class to train a motion progress regression step later. The ground truth for the latter training is estimated from the start and end labels assigned to structural motions and normalized between 0 and 1. In a healthy heart, these are likely to be concurrent with the end-systole and end-diastole frames but there exist slight differences due to motion of the fetus and of the probe. First, it is important to ascertain the presence or absence of the heart. For this, Model-1 was trained on a 35000 frame dataset consisting of annotated viewing plane classes (4C, 3V, LVOT) and background frames. Next, Model-2 was trained to detect the five anatomical structures in the 4C view videos. Such a detection can be treated as a multi-class classification problem for different patches from the image. These patches are treated as individual labelled images that can be fed to a convolutional architecture capable of a 6-way classification (five anatomical structures and background). This training resulted in a model capable of resolving input regions into one of the structure classes. CNNs used have 13 convolutional layers with ReLU activation and 3x3 filters for fine structure aggregation, and 2x2 maxpooling after 2nd, 4th, 7th, 10th and 13th convolutional layers, followed by 2 fully-connected layers of 1024 and 512 units. The last fully-connected layer for Model-1 has 4 units linked to a 4-way softmax classification for visibility and viewing plane estimation. The penultimate fully-connected layer of Model-2 is used to feed spatial information per frame to LSTM modules for accounting temporal information across multiple frames in final detection. The LSTM stages give way to a 6-way softmax for structure classification, and two regression L2 losses, one for motion phase estimation and other for bounding box localization refinement. This is the multi-task convolutional-LSTM with classification and regression nodes for temporal detection and local motion estimation. The latter task of estimating local anatomical motion progress is treated as a local phase detection problem. The phase regression is trained on a quantity defined to be the ratio of relative frame location from the nearest systolic frame to the number of frames between the end-systole to end diastole interval. This value is computed for every structure per frame. While it may appear trivial considering that it is analogous to the cardiac phase itself, such an equivalence holds true only for healthy hearts but digresses in case of wall-motion or structural abnormalities [9] and the difference in local and global motion progress or phase serves as a vital symptomatic marker for disease. Also, the Model-2 is pre-trained on anatomical structure videos on a pure classification stage to pre-condition it with parameters that would allow optimized label assignment for input region proposals. While this may seem redundant given that a multi-task training is performed anyway with both classification, detection, bounding box regression and local estimation phase, the pre-training is found to allow for better convergence and minimize overfitting due to the complexity of the procedure involved.

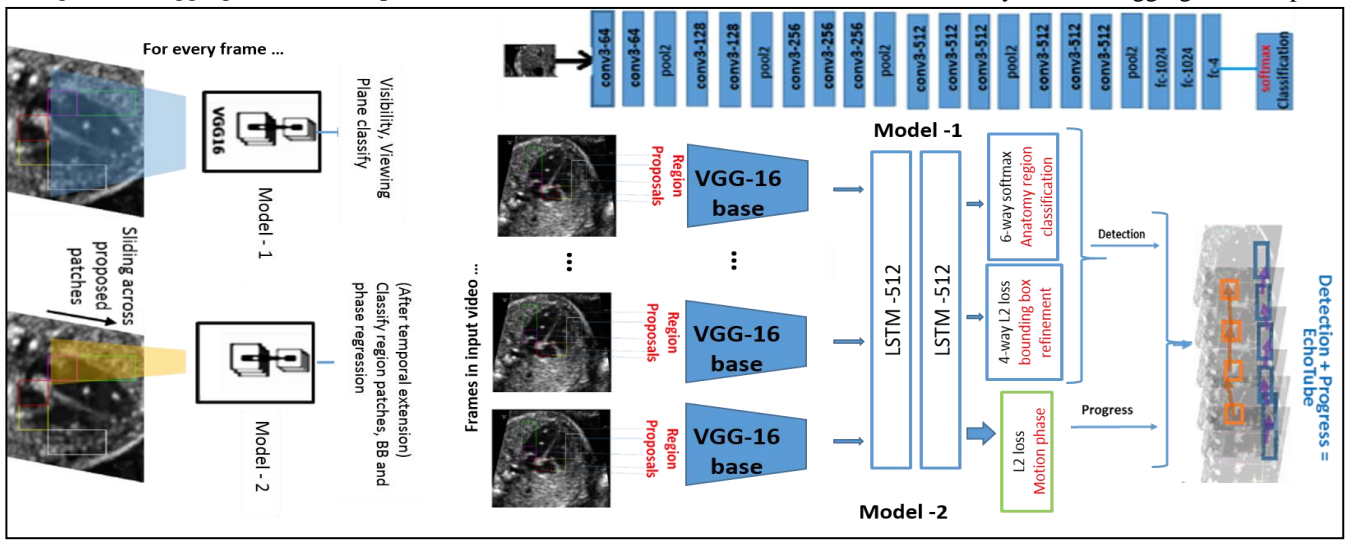

## 4. EXPERIMENTS

91 routine fetal echocardiography videos from 12 healthy subjects of gestational ages ranging from 20 and 35 weeks were available. Each video was between 2 and 10 seconds and a frame rate between 25 and 76 frames per second (39556 frames in total). Videos from 10 subjects were used for training, and 2 for test. For training, we split videos of the 4C view into frames and apply data augmentation by performing an up-down and a top-bottom flip, and a rotational augmentation using angular steps of 10 degrees per frame, excluding for flips. This was applied to patches for five different anatomies, which are extracted in 50x50, and 50x100 pixel resolutions from the frames and used to pre-train Model-2. The original frames are used for the visibility and view classification network training (Model-1). Next, we chose 32 frames at a time in the sequence of their original occurrence and concatenated them into 32-frame videos (the starting frame is an end-diastole frame) and repeated this for augmented data. This resulted in 4760 clips (clips are ignored if sourced from frames adjacent to background or a viewing plane transitions). Both models were trained with a 25-frame mini-batch and a learning rate of 0.01. Model-1 is trained for 200 epochs and Model-2 is trained for 250 epochs during pre-training on individual anatomy clips and fine-tuned along with the regression losses for 100 epochs with the region proposals from the full-frame clips.

## 5. RESULTS

Results for the motion localization task are summarized in Table 1. We briefly discuss the choice of metrics for the 'echo tube' categorization task. One aspect of such a task is the fidelity of temporal anatomy localization. The other aspect is an accurate identification of motion/action bounds. In our task, the former is a result of the efficiency of the object detection step, with a proxy being the average detection accuracy for each anatomical class represented in the echo tube over all the parsed frames, averaged over test instances. The second is a measure of the correctness of the start and end-points.This is presented as a 'frame-difference' metric, averaged over test instances, i.e, we establish the frame number of the source of a patch classified as end of an echo tube and compare with ground truth.

We compared our method in two protocols, i) using only the Model-2 for the localization effort, without including frame-level global features in Protocol–1 ii) with a global context provided by the fully-connected embeddings from the CNN in Model-1 fused with the pre-LSTM fully-connect layers of Model 2 in Protocol-2. Protocol-2 is observed to have a better performance across classes. This reflects in the difference of accuracy in identification of anatomical structures averaged over all the frames in test videos, and in the higher frame differences for identification of motion bounds. Anatomical similarities in some cases, notably valves and ventricle walls, prove to be difficult to accurately distinguish. This is due to a remarkable similarity of appearance in the ultrasound videos used for the task, among other factors like the crop-sizes of the regions used for training and the contextual information encoded, lack of information about heart orientation and the influence of speckle, shadows and enhancements. For motion localization tasks, the frame difference metric for motion patterns of the *formane ovule* and the valves are much lower when averaged for test cases, but the same is not true for the ventricles. This is attributed to the relative inefficiency of using higher-level feature representations to track progress through a regression-based loss when the start and end morphologies are not particularly distinct over adjacent frames. A solution to this might be to use secondary sources of motion patterns, for instance, speckle-tracking. It is worth noting that the efficiency of motion/action detection as modelled using an overall average of per-video framewise average detection accuracies is not a particularly reliable localization performance indicator. This is because object detection is generally more accurate for well-defined patterns and localization relies on frame-to-frame distinction for similar spatial locations. The two factors may not be necessarily correlated over long-interval video sequences despite the physiologically imposed periodicity. In Model-2, the spatial context helps with structure localization and a secondary temporal motion information is introduced through global motion patterns averaged through the recurrent LSTM steps. This is physiologically compatible as local anatomical motion is influenced by overall cardiac cycles. The method does relatively well for *formane ovule* motion categorization. This is an important fetal heart biomarker as it is the opening connecting the left and right atrium and plays a key role in channelizing oxygenated blood into the developing fetus. The motion of ventricle walls are localized as well and their quantified extents over lengths relative to heartbeat intervals serve as vital clinical indicators for conditions of the heart and is an important measure of cardiac function.

To conclude, we presented a deep learning pipeline for quantifying motion patterns for multiple anatomical landmarks in the fetal heart and integrated the automated detection and motion progress which can form the basis for the analysis of diagnostically relevant functional motions in fetal echocardiography videos, as tested on real world clinical data. Though motivated by fetal echocardiography analysis, this method is transferrable to motion analysis in other ultrasound tasks (or to other modalities with video data) and does not require optical flow information.

**Table 1.** Motion/Action detection and localization performances.

| Anatomy | Adapted 2-stream Net [4] | | Echo Tube (ours) | |
|---|---|---|---|---|
| | *Accuracy(%)* | *Bounds aFD* | *Accuracy(%)* | *Bounds aFD* |
| Foramen ovule | 0.9375 | 2.5 | 0.9415 | 2.5 |
| Mitral valve | 0.7185 | 2.3 | 0.7750 | 2.2 |
| Tricuspid valve | 0.7300 | 3.2 | 0.7424 | 3.3 |
| LV wall | 0.8232 | 5.8 | 0.8245 | 4.6 |
| RV wall | 0.8016 | 4.2 | 0.8320 | 4.1 |
| Overall | 0.8022 | 3.6 | 0.8231 | 3.34 |